\documentclass{article}%
\usepackage{graphicx,verbatim}
\usepackage[inkscapelatex=false]{svg}
\usepackage{multicol,multirow}
\usepackage{amsmath,amssymb,amsfonts}
\usepackage{mathrsfs}
\usepackage{amsthm}
\usepackage{rotating}
\usepackage{appendix}
\usepackage[backend=biber,sorting=none]{biblatex}
\bibliography{refs}
\usepackage{ifpdf}
\usepackage[T1]{fontenc}
\usepackage{newtxtext}
\usepackage{newtxmath}
\usepackage{textcomp}
\usepackage{xcolor}
\usepackage{lipsum}
\usepackage[colorlinks,allcolors=blue]{hyperref}
\usepackage{tikz}

\usepackage{caption}
\usepackage[export]{adjustbox}
\theoremstyle{definition}

\numberwithin{equation}{section}
\usepackage{newtxmath}
\DeclareMathAlphabet{\mathpzc}{T1}{pzc}{m}{it}
\usepackage{booktabs}
\usepackage{authblk}

\usepackage{geometry}
\usepackage[none]{hyphenat}
\title{FoilDiff: A Hybrid Diffusion Transformer Model for Airfoil Flow Field Prediction}

\author[1]{Kenechukwu Ogbuagu}
\author[1]{Sepehr Maleki}
\author[1,2]{Giuseppe Bruni}
\author[1,2]{Senthil Krishnababu}

\affil[1]{Lincoln AI Lab, School of Engineering and Physical Sciences, University of Lincoln, United Kingdom}
\affil[2]{Siemens Energy, Lincoln, United Kingdom}
\affil[*]{Corresponding author: \texttt{kogbuagu@lincoln.ac.uk}}

\date{}

\begin{document}
\maketitle

\abstract{
\noindent The accurate prediction of flow fields around airfoils is crucial for aerodynamic design and optimisation. Computational Fluid Dynamics (CFD) models are effective but computationally expensive. This has inspired the development of surrogate models to enable quicker predictions. These surrogate models can be based on deep learning architectures, such as Convolutional Neural Networks (CNNs), Graph Neural Networks (GNNs), and Diffusion Models (DMs). Diffusion models have shown significant promise in predicting complex flow fields. In this work, we propose FoilDiff, a diffusion-based surrogate model with a hybrid-backbone denoising network. This hybrid design combines the power of convolutional feature extraction and transformer-based global attention to generate more adaptable and accurate representations of flow structures. FoilDiff takes advantage of Denoising Diffusion Implicit Model (DDIM) sampling to optimise the efficiency of the sampling process at no additional cost to model generalisation. We use encoded representations of Reynolds number, angle of attack, and airfoil geometry to define the input space for generalisation across a wide range of aerodynamic conditions. When evaluated against state-of-the-art models, FoilDiff shows significant performance improvements, with mean prediction errors reduced by up to 85\% on the same datasets. These results have demonstrated that FoilDiff can provide both more accurate predictions and better-calibrated predictive uncertainty than existing diffusion-based models.}

\section[Introduction]{Introduction}
The accurate and efficient prediction of flow fields across airfoils is crucial for advancing aerodynamic design methodologies of many engineering systems. Computational Fluid Dynamics (CFD) techniques, particularly those based on Reynolds-averaged Navier-Stokes (RANS) equations, are currently used to model flow behaviour \cite{5MATYUSHENKO201715}. However, the numerical complexity of these equations necessitates computationally expensive simulations, making them a bottleneck in iterative design and optimisation processes. As a result, surrogate modelling approaches have gained traction as an effective means to balance computational efficiency and predictive accuracy. Deep learning approaches have demonstrated significant promise in approximating RANS solutions across a wide range of flow conditions \cite{Bruni2023,Thuerey2018,GuoCNN}. Diffusion models are a class of generative models that have been used more recently in flow field predictions, as an alternative to more established deep learning surrogate modelling methodologies, such as Convolutional Neural Networks (CNNs) and standalone U-Net models \cite{gefan2023denoisingdiffusionmodelfluid, hu2024generativepredictionflowfield, Liu2023}. Diffusion models for airfoil flow field prediction, while demonstrating strong generative capability, often require exceptionally lengthy training runs in the tens of millions of iterations or the training of additional models \cite{Liu2023, xiang2024aeroditdiffusiontransformersreynoldsaveraged}. This substantial computational demand presents a major bottleneck, underscoring the need for more efficient training strategies or model architectures to make diffusion-based surrogates more practically viable for aerodynamic design workflows.\medskip

\noindent CNNs have played significant roles in flow field prediction due to their ability to efficiently extract spatial features \cite{ wang2024recentadvancesmachinelearning, ronneberger2015unetconvolutionalnetworksbiomedical,SekarCNN}. CNNs have been used for real-time modelling of non-uniform steady laminar flow, achieving velocity predictions four orders of magnitude faster than CFD solvers with a minor trade-off in accuracy \cite{GuoCNN}. Similarly, they have also been utilized to extract geometric parameters from airfoil shapes, which along with the Reynolds number and angle of attack, were fed into a Multilayer Perceptron (MLP) to approximate flow fields \cite{SekarCNN}. Advancements with encoder-decoder architectures, such as the CNN-FOIL \cite{Duru2022}, have enabled efficient transonic flow field predictions by enhancing shared feature extraction across multiple aerodynamic parameters. Furthermore, the integration of attention mechanisms and U-Net architectures has significantly improved information capture and predictive accuracy in airfoil flow field models \cite{Wang2024, Zuo2024,Bruni2024}.\medskip

\noindent The growing success of deep learning in fluid dynamics has been further propelled by the rise of generative models, which have achieved remarkable results in various applications, including image synthesis \cite{dhariwal2021diffusionmodelsbeatgans}, material design \cite{fu2022materialstransformerslanguagemodels}, and turbulence generation \cite{gao2023bayesianconditionaldiffusionmodels}. In the context of CFD, Generative Adversarial Networks (GAN) models have been applied to turbulence field resolution \cite{kimpredictioncontroltwodimensionaldecaying} and flow field prediction using Variational Autoencoders (VAEs) \cite{ranade2021composableautoencoderbasediterativealgorithm}. An early VAE model for airfoil flow field prediction \cite{Wang2021}, learned a compact mapping from airfoil geometry and flow conditions to full-field pressure and velocity distributions. The model reproduced high quality results at a fraction of the computational cost but tended to smooth sharp gradients near shocks and trailing edges, a common limitation of VAEs with Gaussian latent spaces. To improve generalisation under off-design conditions,  a prior Variational Autoencoder (pVAE) was proposed \cite{Yang2022}. This incorporated a cruise-condition flow field as a latent prior and introduced physics-based loss terms to enforce aerodynamic force balance and mass conservation. To extend this, GAN models adopted an adversarial approach through a data-augmented conditional GAN (daGAN) combining a U-Net-based generator and a discriminator to sharpen flow reconstructions \cite{WU2022}. Data augmentation improved robustness, and the adversarial objective mitigated over-smoothing, yielding clearer shocks and wake features. Despite its success, daGAN still faced typical GAN challenges such as training instability. Diffusion models have emerged as a powerful alternative offering stable training and enhanced mode coverage compared to GANs \cite{liu2021fasterstabilizedgantraining, abaidi2024exploring, dhariwal2021diffusionmodelsbeatgans,liu2023combatingmodecollapsegans}.  \medskip

\noindent Recent studies have demonstrated the effectiveness of denoising diffusion models over standalone CNNs and U-Nets in capturing complex fluid dynamics \cite{gefan2023denoisingdiffusionmodelfluid, SHU2023111972, hu2024generativepredictionflowfield,Liu2023}. Currently, diffusion models applied to the task of airfoil flow field prediction rely on attention augmented U-Nets or standard diffusion transformers as the model backbones \cite{Liu2023,xiang2024aeroditdiffusiontransformersreynoldsaveraged}. Each approach comes with certain strengths and limitations. U-Nets are effective at capturing local dependencies and features but struggle with long-range dependencies \cite{rombach2022highresolutionimagesynthesislatent}.  Attention-augmented U-Nets omit the feed-forward sublayers that are fundamental to standard Transformer blocks, rather using the attention gates to compute attention coefficients by comparing encoder feature maps with gating signals from the decoder \cite{Oktay2018AttentionUNet}. As a result, while they are capable of identifying where long-range interactions should occur, they often lack the depth necessary to encode these dependencies into globally coherent representations of fluid dynamics. This is the architecture applied by the state-of-the-art model Aifnet \cite{Liu2023}. This model demonstrated superior accuracy over standalone UNets in modelling the epistemic uncertainty present in CFD simulations of highly turbulent flows. Despite its strong predictive accuracy, the model’s reliance on millions of training iterations highlights inefficiencies in its learning process, and its generalisation capability remains suboptimal. In contrast, transformer-based diffusion models, such as Diffusion Transformers (DiT) \cite{peebles2023scalablediffusionmodelstransformers}, have demonstrated improved generalisation capabilities over UNet-based diffusion models, relying on the global modelling ability of Vision Transformers (ViTs) \cite{dosovitskiy2021image}, to yield high-fidelity predictions. DiTs for airfoil flow field prediction such as AeroDiT \cite{xiang2024aeroditdiffusiontransformersreynoldsaveraged} introduce separated encoders and decoders to encode inputs into latent space, this approach involves the training of three models, two VAEs and one transformer all trained separately with the trained VAEs used in the transformer training. This multi-stage training process introduces additional computational overhead and tuning complexity. Additionally, the separated encoders and decoders stand to benefit from U-Net-style connections at each resolution layer, to provide additional control and fine tuning over the encoding and decoding to and from the latent space \cite{VAEskips}. \medskip

\noindent To address these limitations, we introduce FoilDiff. FoilDiff combines the benefits of a connected encoder-decoder with a latent transformer within one training process. Additionally, we provide deep conditioning to the latent bottleneck representation to enable more accurate modelling. Finally, FoilDiff employs the Denoising Diffusion Implicit Model (DDIM) accelerated sampling strategy to enhance computational efficiency while maintaining predictive accuracy \cite{song2022denoisingdiffusionimplicitmodels}. To evaluate FoilDiff, we assess its ability to predict steady-state flow fields around across a range of airfoils, Reynolds numbers and angles of attack. We compare performance against state-of-the-art diffusion models with U-Net and transformer backbones. FoilDiff outperforms these baselines in both predictive accuracy and uncertainty modelling.  As some state-of-the-art models differ substantially in scale, rendering direct comparison inappropriate, we further conduct ablation studies to isolate the contributions of each architectural enhancement, highlighting their role in improving model efficiency and expressiveness.\medskip

\noindent The rest of this article is structured as follows: Section 2 provides an overview of diffusion models, detailing the underlying architecture and enhanced sampling strategies. Section 3 outlines the proposed  model architecture, training and inference configurations. Section 4 presents experimental analysis, comparing the performance of our approach to existing models. Finally, Section 5 concludes the paper with key findings and potential directions for future research.\medskip

\section{Preliminary Background on Diffusion Models}
Diffusion models are inspired by the principles of non-equilibrium thermodynamics \cite{pmlr-v37-sohl-dickstein15}. These models operate in a two-process framework: a forward process where noise is added to the data in a gradual process, and a reverse process, in which the model learns how to generate the original data from the noisy input. 

\subsection{Denoising Diffusion Probabilistic Models (DDPMs)}

DDPMs learn to reverse the forward process \( q(\mathbf{x}_t | \mathbf{x}_{t-1}) \) with the conditional distribution \( p_{\theta}(\mathbf{x}_{t-1} | \mathbf{x}_t) \), an estimate of \( q(\mathbf{x}_{t-1} | \mathbf{x}_t) \) \cite{Ho2020}.

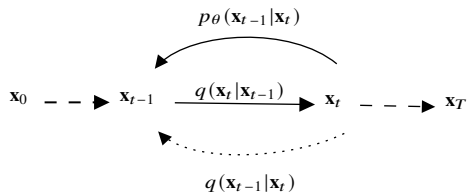
\begin{figure}[h!]
\centering
\tikzset{every picture/.style={line width=0.5pt}}  
 
\begin{tikzpicture}[x=0.75pt,y=0.75pt,yscale=-1.25,xscale=1.25]
\draw  [dash pattern={on 0.84pt off 2.51pt}]  (290.67,167) .. controls (271.67,176.5) and (233.41,179.83) .. (218.49,168.14) ;
\draw [shift={(216.33,166.17)}, rotate = 47.12] [fill={rgb, 255:red, 0; green, 0; blue, 0 }  ][line width=0.08]  [draw opacity=0] (5.36,-2.57) -- (0,0) -- (5.36,2.57) -- cycle    ;
\draw    (287.67,139) .. controls (269.3,123.8) and (230.78,125.75) .. (216.97,138.84) ;
\draw [shift={(215,141)}, rotate = 307.87] [fill={rgb, 255:red, 0; green, 0; blue, 0 }  ][line width=0.08]  [draw opacity=0] (5.36,-2.57) -- (0,0) -- (5.36,2.57) -- cycle    ;
\draw (231,115.4) node [anchor=north west][inner sep=0.75pt]  [font=\scriptsize]  {$p_{\theta }( \mathbf{x}_{t}{}_{-1} |\mathbf{x}_{t})$};
\draw (234,182.4) node [anchor=north west][inner sep=0.75pt]  [font=\scriptsize]  {$q( \mathbf{x}_{t}{}_{-1} |\mathbf{x}_{t})$};
\draw (230,145.4) node [anchor=north west][inner sep=0.75pt]  [font=\scriptsize]  {$q( \mathbf{x}_{t} |\mathbf{x}_{t-1})$};
% Text Node
\draw (156,149.4) node [anchor=north west][inner sep=0.75pt]  [font=\scriptsize]  {$\mathbf{x}_{0}$};
\draw (330,151.4) node [anchor=north west][inner sep=0.75pt]  [font=\scriptsize]  {$\mathbf{x}_{T}$};
\draw (282,150.4) node [anchor=north west][inner sep=0.75pt]  [font=\scriptsize]  {$\mathbf{x}_{t}$};
\draw (200,149.4) node [anchor=north west][inner sep=0.75pt]  [font=\scriptsize]  {$\mathbf{x}_{t-1}$};
\draw    (223,154.67) -- (276,155.35) ;
\draw [shift={(279,155.38)}, rotate = 180.73] [fill={rgb, 255:red, 0; green, 0; blue, 0 }  ][line width=0.08]  [draw opacity=0] (6.25,-3) -- (0,0) -- (6.25,3) -- cycle    ;
\draw  [dash pattern={on 4.5pt off 4.5pt}]  (297,155.69) -- (324,156.24) ;
\draw [shift={(327,156.3)}, rotate = 181.18] [fill={rgb, 255:red, 0; green, 0; blue, 0 }  ][line width=0.08]  [draw opacity=0] (5.36,-2.57) -- (0,0) -- (5.36,2.57) -- cycle    ;
\draw [line width=0.75]  [dash pattern={on 4.5pt off 4.5pt}]  (171,154.5) -- (194,154.5) ;
\draw [shift={(197,154.5)}, rotate = 180] [fill={rgb, 255:red, 0; green, 0; blue, 0 }  ][line width=0.08]  [draw opacity=0] (6.25,-3) -- (0,0) -- (6.25,3) -- cycle    ;
\end{tikzpicture}{\caption{Graphical model of Denoising Diffusion Probabilistic Models}\label{DIFFPROC}}

\end{figure}

\noindent In the forward process, noise is progressively added to the data vector \(\mathbf{x}_0\) across a number of time steps \(T\) , transitioning from 
\( \mathbf{x}_0 \) to \( \mathbf{x}_1 \) to \( \dots \) to \( \mathbf{x}_T \). The intermediate states $\mathbf{x}_t$ can then be expressed as:
\begin{equation}
\mathbf{x}_{t} = \sqrt{\overline{\alpha}_t}\mathbf{x}_{0} + \sqrt{1-\overline{\alpha}_t} \boldsymbol{\epsilon}
\label{eq:ddpm_form}
\end{equation}
\noindent a linear combination of $\mathbf{x}_0$ and a noise variable $\boldsymbol{\epsilon}$, where $\overline{\alpha}_t$ denotes the cumulative product of the variance schedule \cite{Ho2020} 

\begin{equation}
q(\mathbf{x}_{t} | \mathbf{x}_{0}) = \mathcal{N}(\mathbf{x}_{t}; \sqrt{\overline{\alpha}_t}{x_0}, (1-\overline{\alpha}_t)I),
\label{eq:diff_mean}
\end{equation}

\noindent The reverse process then begins with a noisy input and iteratively generates less noisy samples, continuing until \( \mathbf{x}_0 \) is reached from \( \mathbf{x}_T \). To achieve this goal, a network with parameters \( \theta \) is used to model each step of the reverse process, \begin{equation}
p_{\theta}(\mathbf{x}_{t-1} | \mathbf{x}_t) = \mathcal{N}(\mathbf{x}_{t-1}; \mu_{\theta}(\mathbf{x}_t; t), \Sigma_{\theta}(\mathbf{x}_t; t)),
\label{eq:rev diff_mean}
\end{equation}
\noindent where \( \mu_{\theta}(\mathbf{x}_t; t) \in \mathbb{R}^d \) and \( \Sigma_{\theta}(\mathbf{x}_t; t) \in \mathbb{R}^{d \times d} \) are the mean and covariance of the distribution at step \( t \), respectively \cite{ghojogh_ghodsi_2024}. The forward process only requires adding noise to the data; it does not involve any learning and remains fixed. As a result, only the reverse process needs to be trained in diffusion models. This distinguishes diffusion models from VAEs \cite{kingma2022autoencodingvariationalbayes,inbookGhoj}. Following formulation in Equation \ref{eq:ddpm_form}, the prediction for \(\mathbf{x}_{t-1}\) based on $\mathbf{x}_t$ can be expressed as a function of $\boldsymbol{\epsilon}_\theta$, the output of the noise predictor neural network parameterised by $\theta$.
\begin{equation}
\mathbf{x}_{t-1} = \frac{1}{\sqrt{\overline{\alpha}_t}}(\mathbf{x}_{t} - \sqrt{1-\overline{\alpha}_t} \boldsymbol{\epsilon}_\theta)
\label{eq:ddpm_rev_form}
\end{equation}
\medskip

\noindent The training objective is then to minimise the discrepancy between the actual noise \( \boldsymbol{\epsilon} \)  and the noise predicted by the model, \( \boldsymbol{\epsilon}_{\theta}(\mathbf{x}_t, t) \). This objective is formalized using the loss function:

\begin{equation}
L_{\theta} = \mathbb{E}_{\mathbf{x}, \boldsymbol{\epsilon} \sim \mathcal{N}(0, I)} \left[ \| \boldsymbol{\epsilon} - \boldsymbol{\epsilon}_{\theta}(\mathbf{x}_t, t) \|^2 \right]
\label{eq:diff_loss}
\end{equation}
\medskip

\noindent By minimizing this loss function, the model, parameterised by \( \theta \), learns to accurately estimate the noise component within a noisy sample \( \mathbf{x}_t \). Consequently, diffusion models trained under this paradigm serve as powerful generative models capable of producing realistic and coherent data samples. 

\subsection{Non-Markovian Inference}
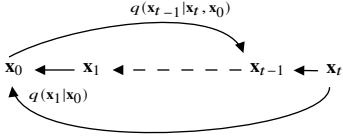
\begin{figure}
\centering
\tikzset{every picture/.style={line width=0.5pt}} 
 %set default line width to 0.75pt        

\begin{tikzpicture}[x=0.75pt,y=0.75pt,yscale=-1,xscale=1]
%uncomment if require: \path (0,257); %set diagram left start at 0, and has height of 257

%Curve Lines [id:da547162887545938] 
\draw    (162.33,146) .. controls (190.8,130.24) and (264.43,118.51) .. (278.09,141.37) ;
\draw [shift={(279.33,144)}, rotate = 250.16] [fill={rgb, 255:red, 0; green, 0; blue, 0 }  ][line width=0.08]  [draw opacity=0] (5.36,-2.57) -- (0,0) -- (5.36,2.57) -- cycle    ;
%Curve Lines [id:da04894648121997114] 
\draw    (322.3,161.14) .. controls (322.3,184.54) and (176.41,197.03) .. (164.48,163.35) ;
\draw [shift={(163.83,160.67)}, rotate = 82.3] [fill={rgb, 255:red, 0; green, 0; blue, 0 }  ][line width=0.08]  [draw opacity=0] (5.36,-2.57) -- (0,0) -- (5.36,2.57) -- cycle    ;

% Text Node
\draw (221.5,116.9) node [anchor=north west][inner sep=0.75pt]  [font=\tiny]  {$q( \mathbf{x}_{t}{}_{-1} |\mathbf{x}_{t} ,\mathbf{x}_{0})$};
% Text Node
\draw (159,147.4) node [anchor=north west][inner sep=0.75pt]  [font=\scriptsize]  {$\mathbf{x}_{0}$};
% Text Node
\draw (319.5,147.9) node [anchor=north west][inner sep=0.75pt]  [font=\scriptsize]  {$\mathbf{x}_{t}$};
% Text Node
\draw (281.33,147.4) node [anchor=north west][inner sep=0.75pt]  [font=\scriptsize]  {$\mathbf{x}_{t-1}$};
% Text Node
\draw (198,147.4) node [anchor=north west][inner sep=0.75pt]  [font=\scriptsize]  {$\mathbf{x}_{1}$};
% Text Node
\draw (171,160.4) node [anchor=north west][inner sep=0.75pt]  [font=\tiny]  {$q( \mathbf{x}_{1} |\mathbf{x}_{0})$};
% Connection
\draw  [dash pattern={on 4.5pt off 4.5pt}]  (278.33,152.5) -- (217,152.5) ;
\draw [shift={(214,152.5)}, rotate = 360] [fill={rgb, 255:red, 0; green, 0; blue, 0 }  ][line width=0.08]  [draw opacity=0] (5.36,-2.57) -- (0,0) -- (5.36,2.57) -- cycle    ;
% Connection
\draw    (195,152.5) -- (178,152.5) ;
\draw [shift={(175,152.5)}, rotate = 360] [fill={rgb, 255:red, 0; green, 0; blue, 0 }  ][line width=0.08]  [draw opacity=0] (5.36,-2.57) -- (0,0) -- (5.36,2.57) -- cycle    ;
% Connection
\draw    (316.5,152.87) -- (307.33,152.73) ;
\draw [shift={(304.33,152.69)}, rotate = 0.84] [fill={rgb, 255:red, 0; green, 0; blue, 0 }  ][line width=0.08]  [draw opacity=0] (5.36,-2.57) -- (0,0) -- (5.36,2.57) -- cycle    ;

\end{tikzpicture}
\caption{Graphical model for non-Markovian inference models}
\label{DDIMPROC}
\end{figure}

The Markovian nature of the conventional diffusion process introduces certain limitations. The step-by-step denoising operation requires a large number of iterative steps to generate high-quality samples. To address these limitations  alternative approaches that relax the Markovian assumption have been proposed, leading to more efficient and expressive generative processes \cite{maleki2025diffusionmodelsmathematicalintroduction}. Non-Markovian inference (see Figure \ref{DDIMPROC}) introduces dependencies across multiple time steps rather than restricting transitions to adjacent steps \cite{song2022denoisingdiffusionimplicitmodels}. This innovation allows for faster sampling since fewer iterations are required to traverse the latent space while still capturing the complexity of the target distribution. This forward process is expressed as:

\begin{equation}
q(\mathbf{x}_{1:T} | \mathbf{x}_0) = \prod_{t=2}^{T} q(\mathbf{x}_{t-1}|\mathbf{x}_t , \mathbf{x}_0),
\label{eq:fwd_ddim}
\end{equation} 
where $q (\mathbf{x}_T|\mathbf{x}_0) = \mathcal{N}( \sqrt{\overline{\alpha}_T}\mathbf{x}_0 ,\ (1-\overline{\alpha}_T)I)$, for all $t>1$. With the mean given by:
\begin{equation}
q(\mathbf{x}_{t-1} | \mathbf{x}_{t}) = \mathcal{N}(\sqrt{\overline{\alpha}_{t-1}}\mathbf{x}_0 +  \sqrt{1 - \overline{\alpha}_{t-1} -\sigma_t^2} \ \ . \  \frac{\mathbf{x}_t - \sqrt{\overline{\alpha}_t}\mathbf{x}_0}{\sqrt{1-\overline{\alpha}_t}}, \sigma_t^2 I),
\label{eq:ddim_mean}
\end{equation}
chosen to ensure that Equation \ref{eq:diff_mean} remains true. One can generate a sample \( \mathbf{x}_{t-1} \) from:

\begin{equation}
\mathbf{x}_{t-1} = \sqrt{\overline{\alpha}_{t-1}} 
{
    \left( \frac{\mathbf{x}_t - \sqrt{1 - \overline{\alpha}_t} \ \boldsymbol{\epsilon}_\theta(\mathbf{x}_t)}{\sqrt{\overline{\alpha}_t}} \right)
}
+ {
    \sqrt{1 - \overline{\alpha}_{t-1} - \sigma_t^2} \cdot \boldsymbol{\epsilon}_\theta(\mathbf{x}_t)
}
+ {\sigma_t \boldsymbol{\epsilon}}
\end{equation}
\medskip

\noindent In this context, \( \boldsymbol{\epsilon} \sim \mathcal{N}(0, I) \) represents standard Gaussian noise, which is independent of \( \mathbf{x}_t \). The selection of \( \sigma_t \) values dictates the nature of the generative process, enabling diverse configurations without necessitating the re-training of the model noise predictor \( \boldsymbol{\epsilon}_\theta \). This approach is formalized as the Denoising Diffusion Implicit Model (DDIM) \cite{song2022denoisingdiffusionimplicitmodels}, an implicit probabilistic model \cite{mohamed2017learningimplicitgenerativemodels}, generating samples deterministically by transitioning from latent variables \( \mathbf{x}_T \) to \( \mathbf{x}_0 \).

\medskip

\noindent DDIMs reduce computational cost by observing that the denoising objective depends only on \( q(\mathbf{x}_t \mid \mathbf{x}_0) \). This allows defining forward processes with fewer steps and enables non-Markovian sampling over a subset rather than the full set \( \mathbf{x}_1, \ldots, \mathbf{x}_T \). As a result, DDIM achieves 10--50$\times$ faster inference without additional training \cite{song2022denoisingdiffusionimplicitmodels}, providing efficiency crucial for practical use.

\subsection{Noise Predictor in Diffusion Models}

The noise predictor, defined in Equation \ref{eq:rev diff_mean}, plays a crucial role in the reverse diffusion process. Diffusion models inherently model conditional distributions \( p(\mathbf{x}_{t-1}|\mathbf{x}_{t}) \), where conditioning enables external information including class labels, captions, or physical parameters to influence the generative process \cite{pan2022arbitrarystyleguidanceenhanced,dhariwal2021diffusionmodelsbeatgans,luo2022understandingdiffusionmodelsunified}. Lyu et al. \cite{lyu2022a} introduced a formulation where physical parameters are integrated as conditions \( \mathbf{c} \) in the noise predictor, resulting in the modified loss function:  

\begin{equation}
L_{NN}(\theta) = \mathbb{E}_{\mathbf{x}, \boldsymbol{\epsilon} \sim \mathcal{N}(0, I)} \left[ \| \boldsymbol{\epsilon} - \boldsymbol{\epsilon}_{\theta}(\mathbf{x}_t,\mathbf{c}, t) \|^2 \right].
\label{eq:diff_loss_cond}
\end{equation}\medskip

\noindent U-Net and Transformer-based neural networks have become the predominant architectures for noise prediction in diffusion models. Originally introduced for biomedical image segmentation \cite{ronneberger2015unetconvolutionalnetworksbiomedical}, the U-Net architecture has since become a popular network architecture for noise prediction in diffusion-based models \cite{rombach2022highresolutionimagesynthesislatent, Ho2020}, due to its ability to reconstruct fine-grained details.  The U-Net structure consists of a contracting path and an expanding path, connected via skip connections. The contracting path reduces spatial resolution while increasing feature depth, enabling hierarchical feature extraction. Conversely, the expansive path reconstructs the output back to its original resolution with skip connections to retain spatial details. This structure ensures a balance between global context understanding and fine-detail preservation, which is crucial for denoising accuracy \cite{prasad2023unravelingtemporaldynamicsunet}. A wide range of conditioning mechanisms can be seamlessly integrated into the U-Net architecture, enabling the model to incorporate external information such as class labels, physical parameters, or auxiliary modalities \cite{yan2022diffusion, yang2024diffusionmodelscomprehensivesurvey,fu2024unveilconditionaldiffusionmodels}.\medskip

\noindent Although convolutional backbones like U-Net remain widely used, transformer-based architectures have emerged as strong alternatives for noise prediction in diffusion models. Diffusion Transformers (DiT) demonstrate superior generative capacity over U-Net-based approaches, especially in large-scale generative tasks \cite{peebles2023scalablediffusionmodelstransformers}. Their core advantage lies in capturing long-range dependencies through self-attention, formulated as:  

\begin{equation}
    \mathrm{Attention}(\mathbf{Q}, \mathbf{K}, \mathbf{V}) = \mathrm{softmax}\left(\frac{\mathbf{Q}\mathbf{K}^\top}{\sqrt{d_{v}}}\right)\mathbf{V},
\label{attention}
\end{equation}  

\noindent where  \(Q, K,V\) represent query, key, and value matrices derived from the encoded conditioning information
and input features, and $d_v$ is the dimension of the value matrix. The self-attention mechanism \cite{Vaswani2017} dynamically weighs different regions of an input feature map, allowing transformers to model complex dependencies beyond local receptive fields. 
\noindent Beyond standard conditioning, integrating attention mechanisms within the U-Net framework enhances diffusion models' adaptability. Inspired by transformer architectures \cite{Vaswani2017}, attention gates in U-Net layers enable flexible multimodal conditioning, improving synthesis quality and domain-specific performance \cite{pmlr-v139-radford21a,raffel2023exploringlimitstransferlearning,hu2024generativepredictionflowfield}. The noise predictor adopted in FoilDiff is designed to integrate the strengths of state-of-the-art architectures, thereby maximising predictive accuracy and generative performance.

\medskip

\section{Model Architecture and Configuration}

\begin{figure}[h!]
    \centering
    \includegraphics[width=0.95\linewidth]{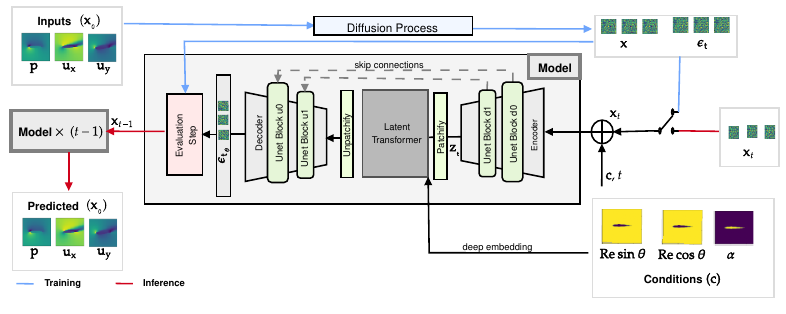}
    \caption{ The FoilDiff architecture, incorporating an encoder, a latent transformer, and a decoder, with U-Net-style skip connections and deep latent embeddings to integrate conditional information effectively}
    \label{fig:FoilDiff-Architecture}
\end{figure}

\noindent The FoilDiff framework integrates a hybrid model architecture, structured dataset, and efficient inference procedures to predict airfoil flow fields across diverse aerodynamic regimes. The overall design, as seen in Figure \ref{fig:FoilDiff-Architecture} leverages a latent transformer, with U-Net-style connected encoder-decoder paths, trained under the canonical denoising diffusion probabilistic modelling paradigm \cite{Ho2020}. Our implementation explores two separate latent transformer architectures and an accelerated DDIM sampling strategy to reduce the number of inference steps required while preserving predictive fidelity \cite{song2022denoisingdiffusionimplicitmodels}. \medskip

\noindent The model is conditioned on encoded physical parameters, including airfoil geometry, Reynolds number, and angle of attack. These parameters are further reinjected into the latent transformer to enable improved generalisation across a wide range of aerodynamic conditions. The combination of hybrid architectural design, domain-aware conditioning, and efficient inference enables FoilDiff to achieve improved accuracy and uncertainty estimation compared to existing U-Net-based baselines. The following subsections detail the model architecture and the methods employed during training and inference.

\subsection{Encoder and Decoder}

\noindent The encoder and decoder components of FoilDiff are designed with U-Net-style skip connections to support a latent transformer. The encoder comprises a sequence of convolutional downsampling blocks, each consisting of two convolutional layers with group normalization and GELU activation, followed by a $2 \times 2$ strided convolution for spatial reduction. Positional encodings and a multi-head self-attention mechanism are added at each resolution level to retain spatial context throughout the hierarchy.\medskip

\noindent The decoder mirrors the encoder structure, employing transposed convolutional blocks for spatial upsampling, interleaved with convolutional layers. To further enhance feature retention, skip connections are incorporated from the encoder to the decoder. These connections facilitate the transfer of spatial information between the contracting and expansive paths, mitigating information loss during downsampling.\medskip

\noindent These skip connections improve gradient flow during training, preventing vanishing gradient issues and aiding in the reconstruction of fine-grained flow structures. Studies \cite{prasad2023unravelingtemporaldynamicsunet, Wang2024} have shown that such architectural enhancements lead to improved convergence rates and more stable training in deep learning-based fluid dynamics simulations.
\medskip

\subsection{Latent Transformer}

\noindent At the core of FoilDiff are latent vision transformer blocks, positioned at the bottleneck of the U-Net-like architecture. These enable better global context modelling and deeper conditioning of the latent by allowing all spatial locations in the latent feature map to attend to one another and the conditions, addressing the limited receptive field of strictly convolutional networks. 

\begin{figure}[h!]
    \centering
    \includegraphics[width=0.8\linewidth]{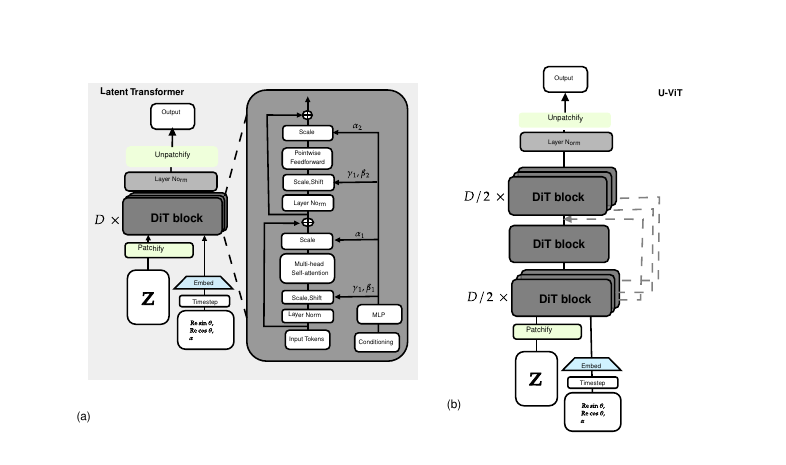}
    \caption{\textbf{Latent transformer architectures.} (a) Standard DiT architecture illustrating the conventional diffusion transformer pipeline \cite{peebles2023scalablediffusionmodelstransformers}; (b) U-ViT architecture showing modifications \cite{bao2023worthwordsvitbackbone}. $Z \in \mathbb{R}^{B \times M}$ denotes the latent representation obtained at the final block of the encoder, where $B$ is the batch size, $M$ the embedding dimension, and $D$ is the total number of blocks}
    \label{fig:Transformer-Archictectures}
\end{figure}
\medskip

\noindent The latent representation, obtained after a series of downsampling operations in the encoder, is first reshaped into a sequence of flattened patches. Each patch is linearly projected into an embedding space and augmented with learnable positional encodings to retain spatial order. These embeddings are then passed through a stack of transformer encoder layers as seen in Figure \ref{fig:Transformer-Archictectures}, each composed of multi-head self-attention and feed-forward sublayers, interleaved with layer normalization and residual connections scaled, reshaped, and modulated by the  physical conditions using AdaLN-Zero \cite{peebles2023scalablediffusionmodelstransformers}. This latent conditioning ensures that the generative process remains grounded in the specified physical scenario.\medskip

\noindent In addition to the standard transformer encoder, we also investigate a latent transformer based on the Unified Vision Transformer (U-ViT) architecture in Figure \ref{fig:Transformer-Archictectures}(b) by applying skip connections between downstream and upstream blocks.   The downstream blocks of the U-ViT require a resolution of the skip connections in the form:
\[
\mathbf{z} = \text{Linear} \left[ \mathbf{z}\, \| \, \mathbf{z}_{\text{skip}} \right],
\]

\noindent the application of a linear layer to the concatenation of the $\mathbf{z}$ and the upstream skip $\mathbf{z}_{skip}$ in order to maintain the dimension of $\mathbf{z}$. Following the final transformer block, the processed sequence is reshaped back into its original spatial dimensions and passed to the decoder for reconstruction.

\subsection{Training and Inference}

\noindent The final stage of the model is the evaluation step, which during training is carried out by minimizing the loss defined in Equation \ref{eq:diff_loss_cond} between the predicted noise and the noise added during the diffusion process. While the FoilDiff framework is compatible with alternatives, such as the $x$ and $v$ parametrisations \cite{salimans2022progressivedistillationfastsampling}, all architectures, including both the baseline U-Net variants and the proposed FoilDiff model, are trained using the canonical denoising diffusion process with the standard $\boldsymbol{\epsilon}$-parametrisation \cite{Ho2020}. This decision is motivated by the need to isolate the effect of architectural modifications, ensure a fair comparison with existing airfoil flow field diffusion models, and preserve the intended focus and scope of this study.\medskip

\noindent During inference, the iterative nature of diffusion models creates a central concern regarding efficiency. We implement a DDIM sampler with strided time step selection strategy defining  a reduced subset with elements \( s \) from the original full set \( \mathbf{x}_1, \dots, \mathbf{x}_T \), where the selection is based on an increment \( n \), in the sequence \cite{song2022denoisingdiffusionimplicitmodels}:

\[
  1, 1+n, 1+2n, \dots, 1+kn,
\]

\noindent where \( n \) is the stride (i.e., the step size between selected time steps), and \( k \) is the number of selected time steps. \( k \) is the largest integer such that

\[
1 + kn \leq T,
\]

\noindent meaning the last selected index \(  \) remains within bounds of trained time steps 
 \(T\). Mathematically, the reduced subset can be expressed as:

\[
\{ \mathbf{x}_s \mid s = 1 + kn, \quad k \in \mathbb{N}_0, \quad s \leq T \},
\]
\medskip

\noindent where \(S\) is the final element of the sampling sequence. This sampling method reduces the number of evaluations needed during inference while preserving generative fidelity. The resulting gains in computational efficiency support the practical application of FoilDiff to high-resolution airfoil flow field prediction.

\section{Experimental Analysis }
\label{sec:data}
This section presents a comprehensive evaluation of FoilDiff, focusing on its predictive accuracy, uncertainty quantification, and computational efficiency in comparison with existing state-of-the-art diffusion-based models. For training and evaluation, we utilize a dataset consisting of flow field simulations across a range of Reynolds numbers and angles of attack from a publicly available repository \cite{TUM-IRepository1}. Each sample in this dataset comprises a structured grid representation of the flow field, including pressure and velocity vectors, alongside the corresponding airfoil geometry \( (\omega) \), Reynolds number \( (Re) \), and angle of attack \( (\phi) \), and the resulting CFD data. To ensure a compact and numerically stable representation, the Reynolds number and angle of attack were transformed into a parametric encoding using \( (Re\cos(\phi)) \) and \( (Re\sin(\phi)) \), respectively. These normalised physical parameters were embedded as additional channels alongside a mask of the airfoil geometry (\( \omega \)) in a three-channel tensor, which conditions the model \cite{Liu2023}. \medskip

\begin{figure}[h!]
    \centering
    
    \includegraphics[width=0.85\linewidth]{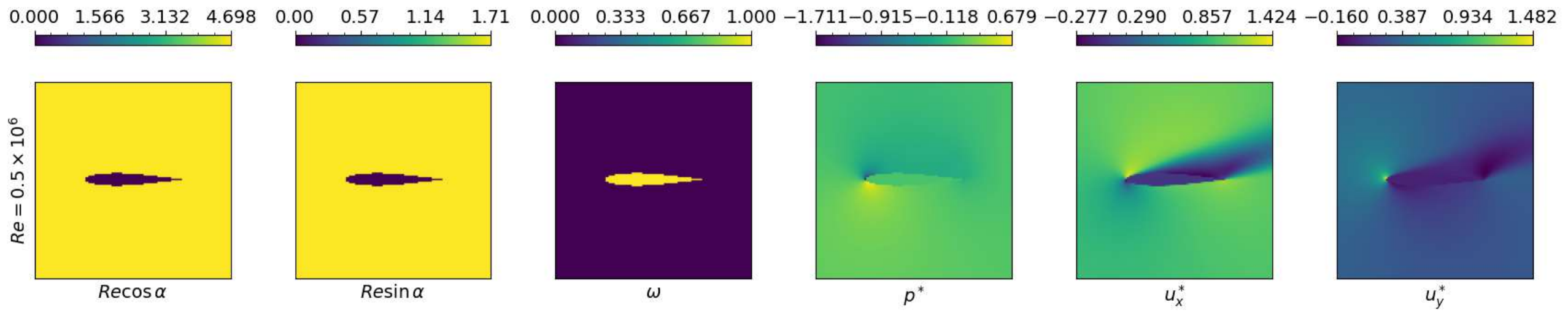}{
    \caption{Structure of the dataset}\label{Ground Truth}}
\end{figure}

\noindent The corresponding CFD case data is  a three-channel tensor with the first channel encoding the dimensionless pressure field:

\begin{equation}
p_i = \frac{p_i - p_{i,\infty}}{|\mathbf{u}_f|^2}
\end{equation}

\noindent and the remaining two channels corresponding to the normalised velocity components:

\begin{equation}
(u_{x,i}, u_{y,i}) = \left( \frac{u_{x,i}}{|\mathbf{u}_f|}, \frac{u_{y,i}}{|\mathbf{u}_f|} \right),
\end{equation}

\noindent where $p_i$ is the normalised pressure at location $i$,   $p_{i,\infty}$ is the reference (freestream) pressure, $|\mathbf{u}_f|$ is the freestream velocity magnitude used for normalization, and $u_{x,i}$ and $u_{y,i}$ are the $x$- and $y$-components of the velocity at $i$. These were concatenated with the conditions, resulting in the six channel \([128\times128]\) tensor shown in Figure \ref{Ground Truth} which is easily downsampled to \([64\times64]\) and  \([32\times32]\) for this study. These normalised representations enhance stability during training and ensure consistency across varying flow conditions.\medskip

\noindent Our experiments are designed to provide a fair and systematic benchmark by following evaluation protocols consistent with prior studies in diffusion-based flow field modelling. We aim to assess both the quantitative performance and the physical fidelity of FoilDiff’s predictions across a range of aerodynamic conditions.\medskip

\subsection{Flow Field Prediction Analysis}

\noindent The flow field prediction task assesses the model’s ability to reconstruct detailed aerodynamic structures, including boundary-layer evolution, surface pressure distributions, and separated flow regions. These structures are key determinants of airfoil performance metrics. FoilDiff is trained on the \([64\times64]\) dataset described in Section~\ref{sec:data}, and tested on a withheld dataset comprising diverse airfoil geometries, Reynolds numbers, and angles of attack. Quantitative performance is measured via the mean squared error (MSE) of the reconstructed fields relative to high-fidelity CFD ground truth, while qualitative analysis examines the spatial coherence and physical realism of the generated flow structures.\medskip

\begin{table}[h!]
    \centering
    \caption{Quantitative evaluation of FoilDiff’s prediction accuracy on selected test cases. Reported metrics represent mean squared error (MSE) between predicted and CFD ground-truth fields.}
    
    \label{tab:predmse}
    \renewcommand{\arraystretch}{1}
    \begin{adjustbox}{max width=0.9\linewidth}
    \begin{tabular}{llccccc}
\toprule
&\multicolumn{3}{c}{Inputs} & \multicolumn{3}{c}{$\mathrm{(MSE)} \times 10^{-3}$}  \\
\cmidrule(lr){2-4} \cmidrule(lr){5-7} 
Case & \textbf{Airfoil} & $(Re)\times10^6$ & \textbf{\(\phi\)} & $\mathbf{p}$ & \textbf{$\mathbf{u_x}$} & \textbf{$\mathbf{u_y}$} \\
\midrule
1& GOE331 & 7.657&-13.08 &0.031 & 0.135 & 0.027 \\
2& CLARYM18&3.618 &6.83 &0.007  &0.042 & 0.020 \\
3& E342&1.875&4.80 &0.006 &0.016 &0.021  \\
\bottomrule
    \end{tabular}
\end{adjustbox}
\end{table}
\medskip

\noindent Table~\ref{tab:predmse} reports the MSE values for pressure (\(p\)) and velocity components (\(u_x, u_y\)) across representative test cases. FoilDiff achieves consistently low errors across all channels, with pressure field predictions exhibiting particularly high accuracy. The velocity components, while slightly more variable, remain within an acceptable tolerance for predictive aerodynamic modelling. These results indicate that the model captures flow patterns with balanced precision.\medskip

 \begin{figure}[h!]
    \centering
    \includegraphics[width=1.3\linewidth]{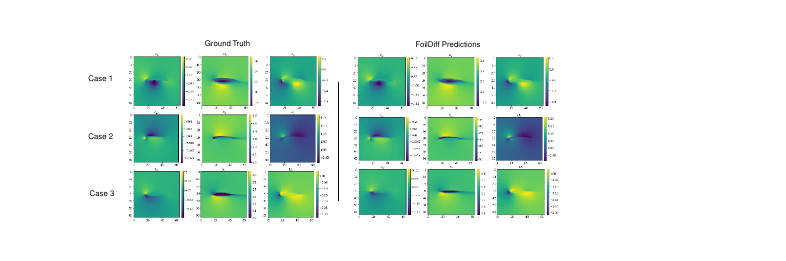}
    \caption{ Comparison of predicted and reference flow fields for representative test cases. The left side displays the ground-truth CFD flow fields, and the right FoilDiff’s reconstructed pressure and velocity fields.
    }
    \label{fig:flowfield}
\end{figure}\medskip

\subsection{Uncertainty Modelling}
\label{se} 
\begin{figure}[h!]
    \centering
    \includegraphics[max width=0.7\linewidth]{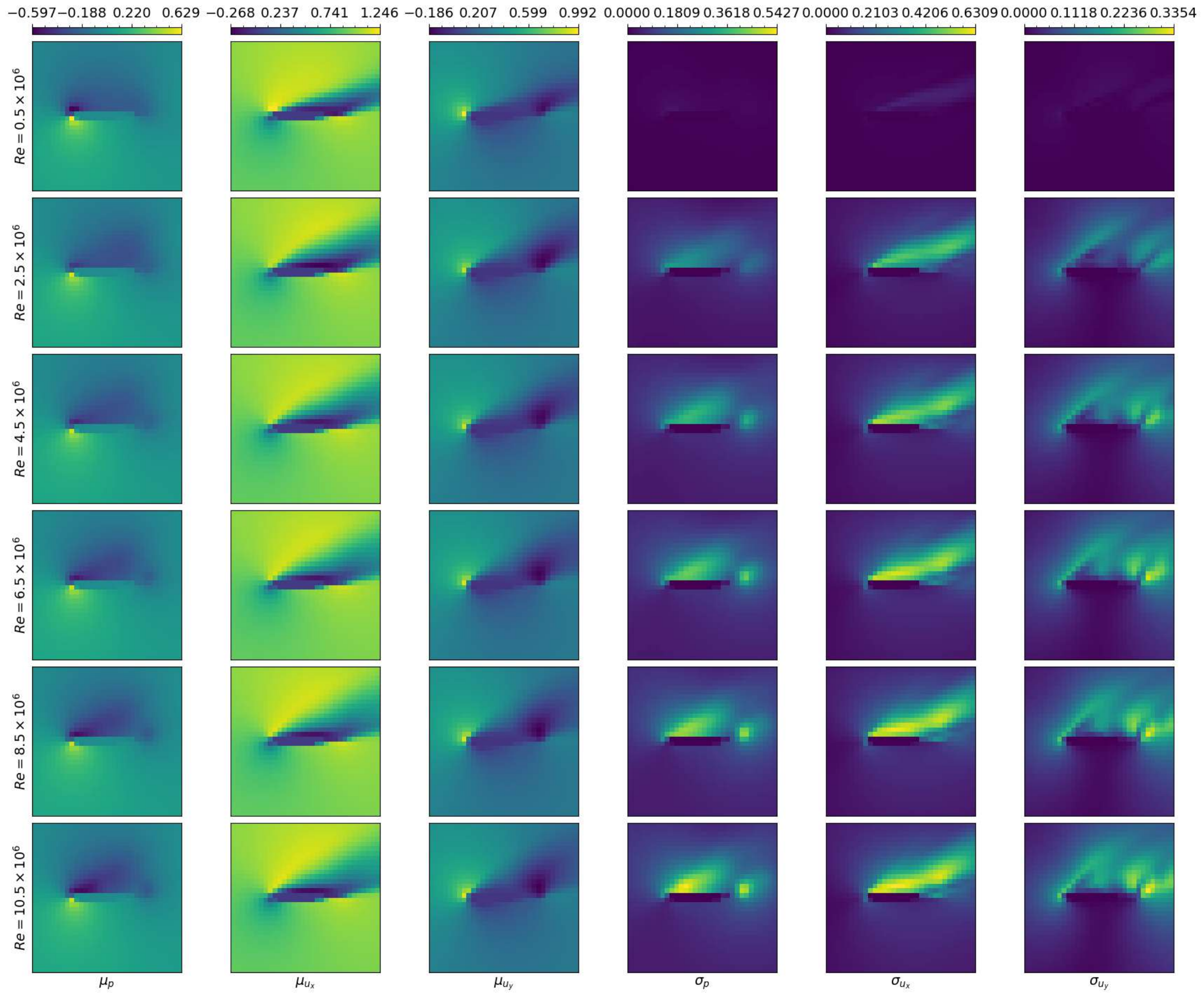}
    \caption{Calculated mean and standard deviation fields from 20 samples of 
    individual cases in the \([32\times32]\) subset}
    \label{fig:Mean fields}
\end{figure}

The aim is to assess the model’s ability to generate high-fidelity flow fields and capture the epistemic uncertainty present in CFD simulations of highly turbulent flows, which becomes aleatoric uncertainty in the data \cite{Liu2023}. We evaluate cases of RAF30 airfoil geometry at $20^\circ $ across a range of Reynolds numbers, as summarized in Table \ref{Testcases}, with cases in regions of higher and lower uncertainty within and beyond the training range. 
For each case in Table \ref{Testcases}, we compute the pointwise mean and standard deviation from 20 simulations under identical input conditions. The resulting fields, shown in Figure \ref{fig:Mean fields}, serve as a reference for both qualitative and quantitative evaluation of predictive uncertainty in our generative framework. Training and inference are performed on a single NVIDIA A100 GPU, and the model performance is measured via the mean squared error (MSE) on both the calculated mean $(MSE_{\mu})$ and standard deviation $(MSE_{\sigma})$ fields of the predicted velocities and pressure fields after 20 samples.

\begin{table}[h!]
\centering
\caption{Case configurations, including Reynolds number and subset assignment}
\label{Testcases}
\renewcommand{\arraystretch}{1.1}
\begin{adjustbox}{max width=0.3\linewidth}
\begin{tabular}{lcc}
\hline
\textbf{Case} & \textbf{Reynolds Number} & \textbf{Subset}\\
\hline
0   & $0.5 \times 10^6$ &Test  \\
1   & $1.5 \times 10^6$ &Training \\
2   & $2.5 \times 10^6$ &Test  \\
3   & $3.5 \times 10^6$ &Training \\
4   & $4.5 \times 10^6$ &Test  \\
5   & $5.5 \times 10^6$ &Training  \\
6   & $6.5 \times 10^6$ &Training \\
7   & $7.5 \times 10^6$ &Test\\
8   & $8.5 \times 10^6$ &Training \\
9   & $9.5 \times 10^6$ & Test \\
10  & $10.5 \times 10^6$ &Test \\
\hline
\end{tabular}
\end{adjustbox}
\end{table}

\noindent Regions of higher epistemic uncertainty are concentrated near the leading and trailing edges, as well as in wake regions where flow separation occurs. To quantify predictive accuracy, we benchmark FoilDiff against a scaled-up state-of-the-art model, Aifnet \cite{Liu2023}. Table~\ref{performance} reports the calculated mean $(MSE_{\mu})$ and standard deviation $(MSE_{\sigma})$ fields of the predicted fields across 20 samples. The uncertainty categories are grouped into lower and higher relative to case 5, which serves as the reference. The test set also included extrapolation cases outside the range of the training set. \medskip

\medskip

\noindent FoilDiff achieves substantial improvements in both interpolation and extrapolation regimes. In interpolation cases, FoilDiff reduces $MSE_\mu$ by 42.0\% in low-uncertainty and 73.3\% in high-uncertainty conditions, while $MSE_\sigma$ decreases by 78.5\% and 75.2\%, respectively. Averaged across all interpolation cases, FoilDiff delivers a 60.5\% reduction in $MSE_\mu$ and a 76.6\% reduction in $MSE_\sigma$. In extrapolation cases, the model reduces $MSE_\mu$ by 77.1\% in low-uncertainty and 84.7\% in high-uncertainty conditions. Correspondingly, $MSE_\sigma$ improves by 95.2\% in high-uncertainty regions, although a modest 8.3\% increase is observed in the low-uncertainty subset. Overall, FoilDiff reduces extrapolation $MSE_\mu$ by 69.6\% and $MSE_\sigma$ by \textbf{88.2\%} across all cases.\medskip

\noindent These results demonstrate that FoilDiff not only provides higher-fidelity mean flow predictions but also offers more reliable uncertainty calibration compared to Aifnet. The gains are particularly pronounced in challenging extrapolation scenarios, underscoring the importance of the hybrid backbone and deep conditioning mechanisms for robust generalisation across unseen flow conditions. The rationale for using a scaled-up version of Aifnet, Aifnet (Extra Large), is discussed in the following section on computational efficiency.

\begin{figure}[h!]
    \centering
    \includegraphics[width= 0.8\textwidth]{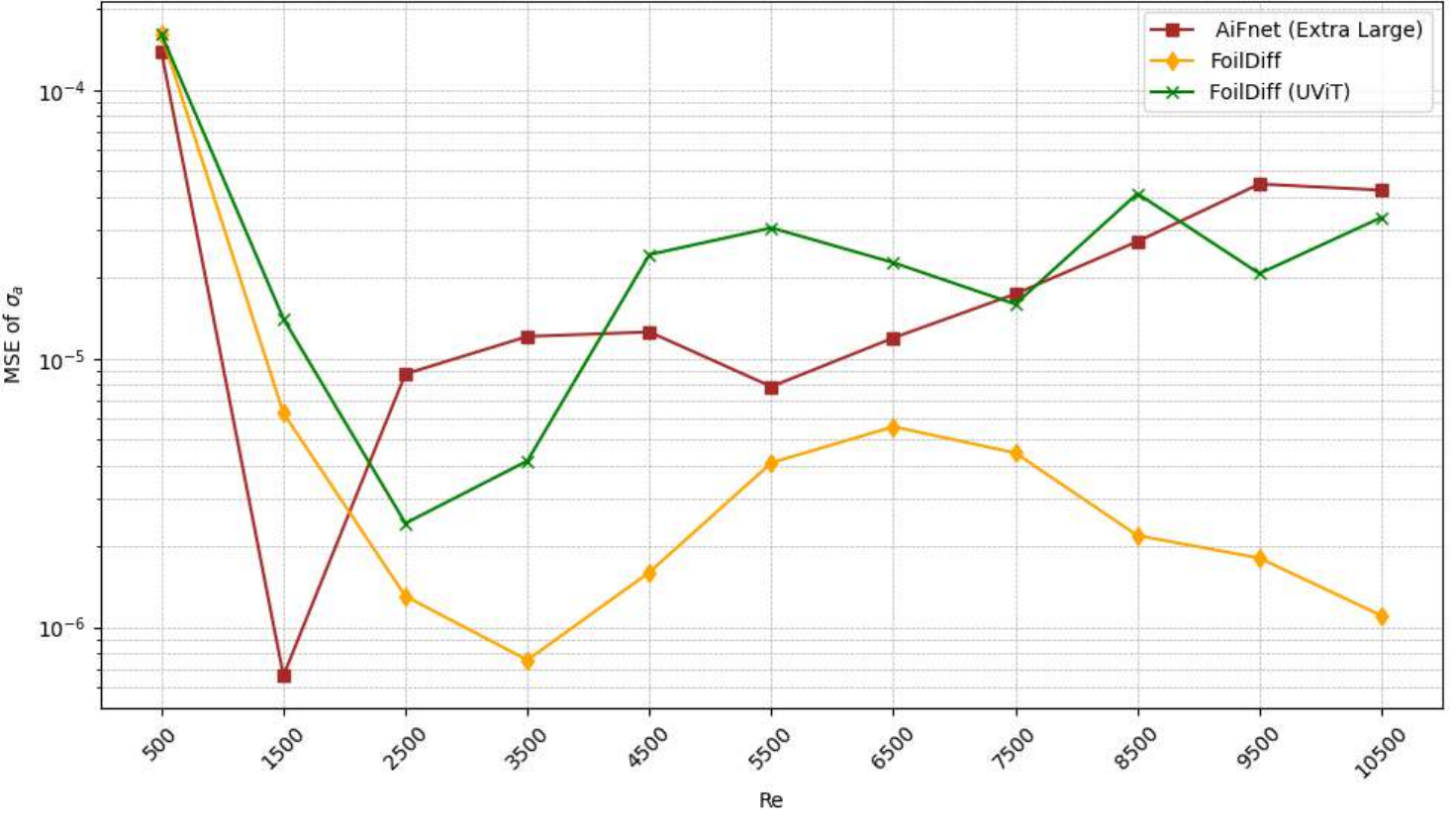}
    \caption{Mean squared error (MSE) comparison of scaled-up Aifnet (Extra Large) and FoilDiff variants across Reynolds numbers after $150k$ training iterations}
    \label{fig:MSE}
\end{figure}\medskip

\begin{table}[h!]
    \centering
    \caption{Average MSE of model predictions on the test dataset, with cases where FoilDiff outperforms the scaled-up state-of-the-art U-Net-based model shown in boldface}
    \label{performance}
    \renewcommand{\arraystretch}{1}
    \begin{adjustbox}{max width=0.9\linewidth}
    \begin{tabular}{llcccc}
\toprule
\multicolumn{2}{c}{} & \multicolumn{2}{c}{($\mathrm{MSE}_{\mu}) \times 10^{-3}$} & \multicolumn{2}{c}{($\mathrm{MSE}_{\sigma}) \times 10^{-3}$} \\
\cmidrule(lr){3-4} \cmidrule(lr){5-6}
\textbf{Dataset Region} & \textbf{Uncertainty Categories} & \textbf{Aifnet} & \textbf{FoilDiff} & \textbf{Aifnet} & \textbf{FoilDiff} \\
\midrule
\multirow{3}{*}{Interpolation region} 
& low $\sigma_y$ cases  & 0.314$\pm$0.043 & \textbf{0.182$\pm$0.024} & 0.079$\pm$0.011 & \textbf{0.017$\pm$0.004} \\
& high $\sigma_y$ cases & 1.005$\pm$0.041 & \textbf{0.268$\pm$0.033} & 0.113$\pm$0.018 & \textbf{0.028$\pm$0.006} \\
& All cases             & 0.532$\pm$0.038 & \textbf{0.210$\pm$0.027} & 0.094$\pm$0.015 & \textbf{0.022$\pm$0.005} \\
\midrule
\multirow{3}{*}{Extrapolation region} 
& low $\sigma_y$ cases  & 1.465$\pm$0.132 & \textbf{0.335$\pm$0.041} & \textbf{1.200$\pm$0.120} & 1.300$\pm$0.150 \\
& high $\sigma_y$ cases & 1.894$\pm$0.169 & \textbf{0.290$\pm$0.055} & 0.273$\pm$0.040 & \textbf{0.013$\pm$0.004} \\
& All cases             & 1.453$\pm$0.127 & \textbf{0.442$\pm$0.049} & 0.450$\pm$0.060 & \textbf{0.053$\pm$0.050} \\
\bottomrule
    \end{tabular}
\end{adjustbox}
\end{table}
\medskip

\subsection{Computational Efficiency}

FoilDiff achieves high predictive accuracy while maintaining a favourable balance of computational cost in both training and inference. Due to computational budget constraints, the models in this study were not trained for the 12.5 million iterations reported in the Aifnet paper \cite{Liu2023}. Instead, all variations of FoilDiff and Aifnet were trained for 150,000 iterations. At this iteration count, FoilDiff not only surpasses the recommended Aifnet configuration for the \([32\times32]\) dataset but also achieves performance superior to that reported in the original Aifnet study \cite{Liu2023}. Requiring over 83 times fewer training iterations with a trade-off in inference time with a model only 10 times larger.\medskip

\begin{figure}[h!]
    \centering
    \includegraphics[width=0.7\linewidth]{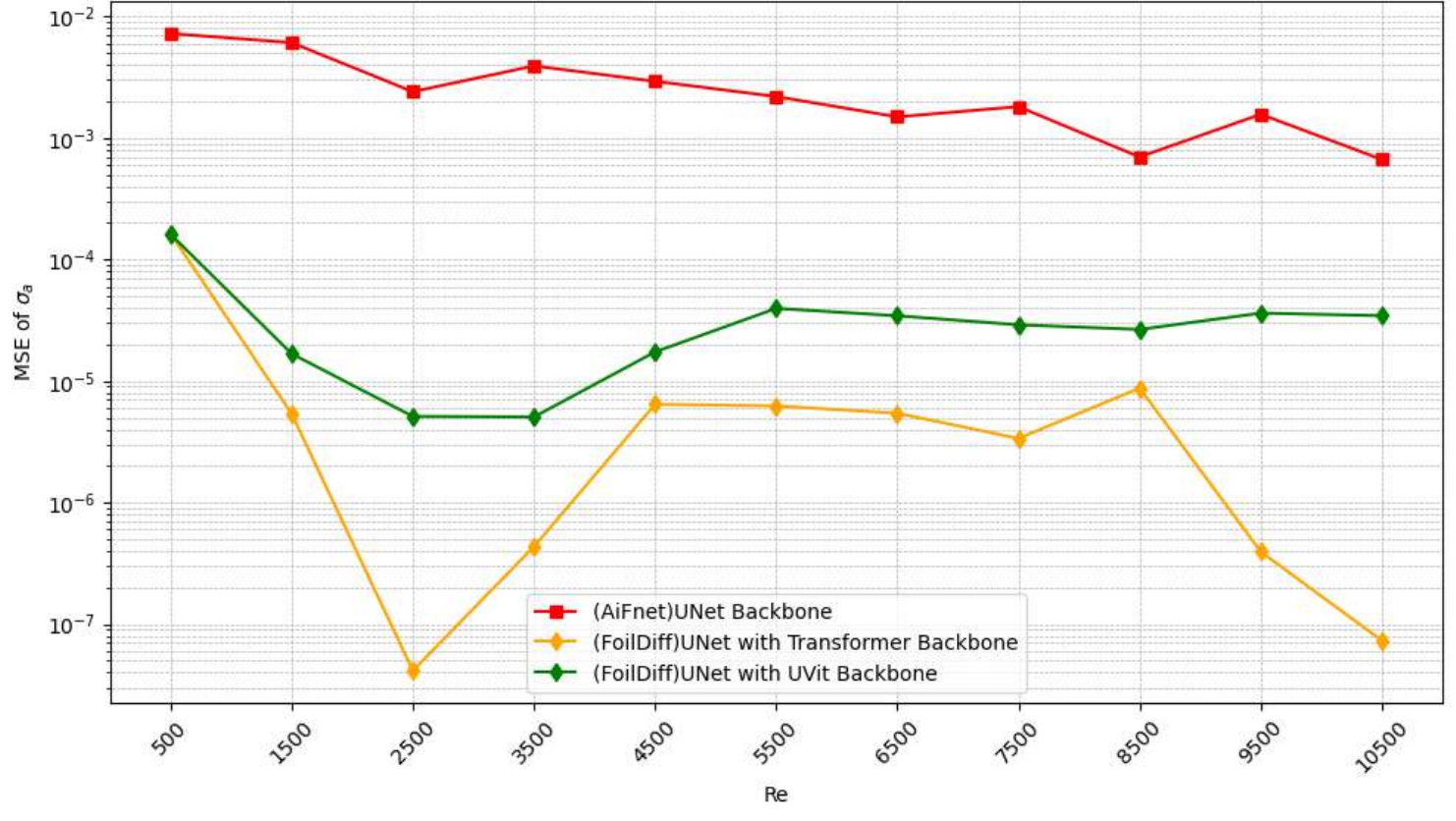}
    \caption{Logarithmic plot of mean squared error (MSE) of the recommended-sized Aifnet model compared with FoilDiff variants after $150k$ training iterations}
    \label{fig:MSE800}
\end{figure}

\noindent While the recommended Aifnet size for the \([32\times32]\) dataset is computationally light, its predictive capacity is limited at 150,000 iterations as shown in Figure \ref{fig:MSE800}. This makes it an insufficient baseline against which to evaluate our hybrid transformer-based approach. To address this, we benchmark against a scaled-up version of Aifnet originally used for the \([128\times128]\) dataset in its paper. The scaled up Aifnet exhibits substantially better predictive accuracy, but FoilDiff still far outperforms this model after the same number of training iterations, as shown in Figure \ref{fig:MSE}.\medskip 

\noindent The efficiency trade-offs are summarized in Table \ref{Rec-cons}. Compared to the scaled-up Aifnet, FoilDiff with a DiT backbone has approximately 38\% more trainable parameters (26.5M vs.\ 19.2M), resulting in 23\% longer training time per run (16 min vs.\ 13 min), and 19\% more inference time per sample (0.032 s vs.\ 0.027 s). These modest overheads are outweighed by the substantial accuracy gains presented in Section \ref{se}, underscoring the effectiveness of the architectural and conditioning choices in FoilDiff. Collectively, these results demonstrate that FoilDiff offers a strong balance of predictive fidelity and computational efficiency, supporting its practical deployment in real-world fluid simulation tasks where both computational resources and time are constrained.\medskip

\medskip 

\begin{table}[h]
    \centering
    \caption{Resource consumption of the model on a single NVIDIA A100 GPU compared with CFD results obtained on an Intel Core i9-13900K processor}
    \label{Rec-cons}
    \renewcommand{\arraystretch}{1.1}
    \begin{adjustbox}{max width=0.7\linewidth}
    \begin{tabular}{lccc}
        \hline
        & \textbf{Number of Parameters} & \textbf{Training time (min) }  & \textbf{Inference time (s)}\\
        \hline
        FoilDiff  &26485145 & 16  & 0.032\\
        FoilDiff (U-ViT)  &39184537 & 25  & 0.043
                 \\
        Aifnet   & 19228565 & 13  & 0.027\\
        Aifnet (recommended) & 1985218 & 5  & 0.018\\
        DiT baseline& 16682896 & 20 & 0.033 \\
        CFD & - & - & 180  
                \\
        \hline
    \end{tabular}
    \end{adjustbox}
\end{table}

\subsection{Ablation Studies}
\noindent To evaluate the individual contributions of FoilDiff’s architectural and algorithmic components, a series of ablation experiments were performed in which specific modules were selectively removed or substituted. The complete FoilDiff model was first compared with a variant in which the latent transformer was replaced by a purely convolutional U-Net mid-block. This modification led to a notable increase in mean squared error (MSE), demonstrating the critical role of the latent transformer and deep conditioning for capturing complex aerodynamic flow structures.\medskip

\noindent A second variant excluded the U-Net-style skip connections between the encoder and decoder, thereby reducing the model to a standard latent diffusion transformer. This omission yielded inferior performance under identical training conditions, confirming the importance of multi-scale feature propagation in preserving local flow details and enhancing reconstruction accuracy.\medskip

\begin{table}[h!]
    \centering
    \caption{Ablation study results showing calculated mean squared error for the mean field $(MSE_{\mu})$ and standard deviation $(MSE_{\sigma})$ on RAF30 airfoil with 20 degrees angle of attack in flow with Reynolds number $7.5 \times 10^6$}
    \label{tab:ablation}
    \resizebox{0.9\textwidth}{!}{%
    \begin{tabular}{lcccccc}
\hline
\textbf{Model Variant} & 
\multicolumn{1}{c}{$MSE_{\mu} \times 10^{-3}$} & 
\multicolumn{1}{c}{Rel. Change} & 
\multicolumn{1}{c}{$MSE_{\sigma} \times 10^{-4}$} & 
\multicolumn{1}{c}{Rel. Change} & 
\multicolumn{1}{c}{Inference Time (s)} & 
\multicolumn{1}{c}{Rel. Change} \\ 
\hline
Full FoilDiff & \textbf{0.526} & -- & 0.481 & -- & 0.032 & -- \\
No Transformer (U-Net mid-block) & 1.612 & +206.5\% & 1.303 & +186.4\% & \textbf{0.025} & \textbf{-21.9\%} \\
Standard DiT (pre-trained vae) & 1.206 & +129.3\% & 0.930 & +104.4\% & 0.033 & +3.1\% \\
FoilDiff no encoder-decoder skips & 0.811 & +54.4\% & 0.809 & +68.1\% & 0.029 & -10\% \\
Full-step DDPM (no accel.) & 0.562 & +6.8\% & \textbf{0.455} &\textbf{-5.7}\%  & 0.184 & +475\% \\
\hline
    \end{tabular}}
\end{table}\medskip

\noindent Contrary to our initial expectation, the U-ViT latent transformer underperformed relative to the standard DiT backbone in FoilDiff. Across the test range in  Figures \ref{fig:MSE} and \ref{fig:MSE800}, we observe consistently higher MSE for the U-ViT variant, despite its larger capacity. One possible reason is that cross-scale information routing, which is effective in convolutional architectures, does not transfer directly to ViT networks. This would require further investigation and tuning of the skip mechanism beyond the scope of this current study. Accordingly, we adopt the standard transformer as the default latent transformer in FoilDiff.\medskip

\noindent Table \ref{tab:ablation} summarizes the results of the ablation study. Overall, the largest accuracy gains are attributable to the hybrid backbone and deep conditioning, while DDIM sampling primarily contributes to efficiency. The combination of all three components yields the best balance between accuracy, generalisation, and computational cost.\medskip

\begin{figure}[h!]
    \centering
    \includegraphics[width=0.95\linewidth]{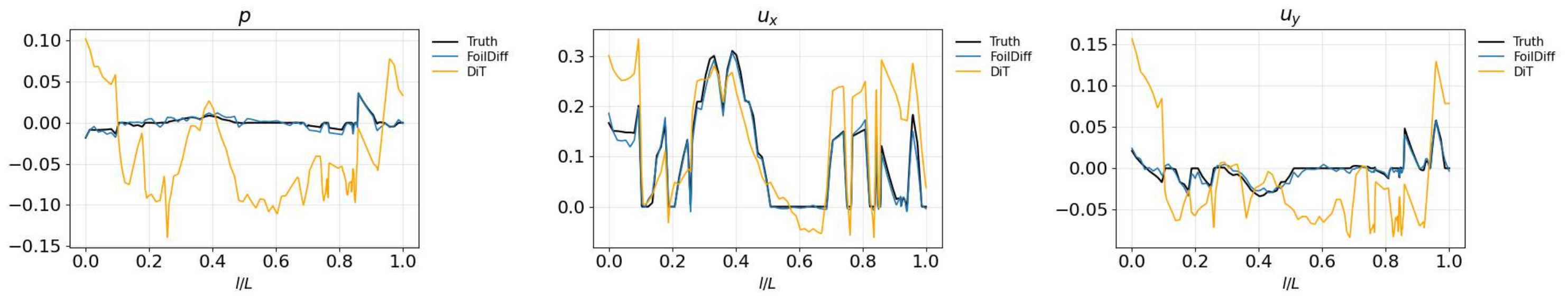}
    \caption{Surface flow quantities predicted by FoilDiff and a DiT-based baseline against the CFD ground truth for E598 airfoil at \( Re = 6.482 \times 10^6\) at an angle of \(-2^\circ\). The three subplots (left to right) show the distributions of the surface pressure coefficient (\(p\)) and the streamwise and normalised vertical velocity components (\(u_x,\, u_y\)) respectively, along the airfoil surface.}
    
    \label{surfacevalues}
\end{figure}\medskip 

\noindent In addition, the DiT configuration was assessed following the canonical procedure outlined in previous diffusion transformer studies, including AeroDiT \cite{peebles2023scalablediffusionmodelstransformers,xiang2024aeroditdiffusiontransformersreynoldsaveraged}. A direct comparison with AeroDiT was not undertaken, as its implementation is not publicly available and reproducing it exactly was beyond our computational budget. Instead, we adopted the reported methodology, which employs pre-trained variational autoencoders, to replace the encoder and decoder components of our model while retaining the latent transformer. This configuration produced poorer results compared with the FoilDiff variant without skip connections, and significantly underperformed relative to the full FoilDiff architecture. Although this approach suggests a more compact model, its limited architectural flexibility and lack of fine-tuning of the latent space to our task-specific features during the encoding and decoding phases resulting in computational, and performance inefficiencies in Table \ref{Rec-cons} and Figure \ref{surfacevalues}. The performance gap is clearly illustrated in Figure \ref{surfacevalues}, here, the horizontal axis represents the \textit{normalised arc length} \(l/L\), defined as the non-dimensional cumulative distance measured along the airfoil surface from the leading edge (\(l=0\)) to the trailing edge (\(l/L=1\)), where \(L\) denotes the total surface length. FoilDiff closely follows the ground truth for all quantities, accurately reproducing the suction-side pressure recovery and the smooth velocity variation around both the upper and lower surfaces. By contrast, the DiT baseline exhibits far greater deviation. These trends highlight FoilDiff’s superior capability to reconstruct coherent and physically consistent flow field and boundary-layer behaviour along the airfoil surface.\medskip

\section{Conclusion}

\noindent This paper introduced FoilDiff, a hybrid diffusion model for predicting airfoil flow fields with improved accuracy and uncertainty estimation. By integrating U-Net-style skip connections in the encoder and decoder of a latent transformer, FoilDiff captures both local and global flow features, enabling more accurate predictions across a range of Reynolds numbers and aerodynamic conditions. FoilDiff outperforms existing state-of-the-art diffusion-based models for airfoil flow prediction while requiring significantly fewer training iterations. Our architecture achieves both high predictive accuracy and well-calibrated uncertainty estimates, with up to 85\% drop in prediction mean square error when directly compared with state-of-the-art models. The connected encoder-decoder for a latent transformer marks a significant step forward in diffusion transformer model design achieving significantly better performance than conventional architectures.\medskip

\noindent FoilDiff serves as a proof of concept demonstrating the potential of a hybrid diffusion transformer for airfoil flow field prediction. However, these achievements alone do not yet establish FoilDiff as a practical tool for real-world fluid simulation. A general challenge to diffusion models is their large parameter counts and the disproportionate computational resources they require for relatively simpler tasks such as 2D airfoil flow field prediction compared to several other deep learning techniques. For diffusion-based frameworks to transition from promising research to practical implementation, further progress is needed in several key areas: improving computational efficiency and advancing integration with physics-informed or hybrid modelling approaches. Only through such developments can the promise of diffusion models be fully realised. \medskip

\noindent Building on this success in the case of a two-dimensional airfoil flow field prediction, future work will extend FoilDiff to more complex flow field configurations. We intend to expand the scope beyond the current limitations of this paper investigating the application of diffusion models to more complex problem spaces including unsteady and three-dimensional flows, the modelling of rotational interactions and airfoil cascades, and multitask systems and aeromechanics. A central objective would be to determine if the scientific value justifies the compute and where the most compute gains could be against CFD techniques, which requires careful accounting of wall-clock training time, energy usage, and memory footprint relative to accuracy gains. Another would be the investigation of physics-informed constraints that enforce conservation laws, symmetries, and boundary conditions; performance-efficient methodologies and development of  hybrid pipelines that couple generative priors with other deep learning models, reduced-order models or classical solvers. Equally important would be rigorous assessment beyond mean error and calibrated uncertainty to include sensitivity to meshing and geometry parametrisation. Our overall goal is to make diffusion models not only high performing but also credibly interpretable, resource-conscious, and practically deployable alternatives to conventional CFD.

\paragraph{Acknowledgments}

\paragraph{Competing Interests}
There are no financial, professional, contractual, or personal relationships or situations that could be perceived to influence the presentation of this work.

\paragraph{Data Availability Statement}
The dataset is available at https://dataserv.ub.tum.de/index.php/s/m1459172

\paragraph{Ethical Standards}
The research meets all ethical guidelines, including adherence to the legal requirements of the study country.

\paragraph{Author Contributions} Writing original draft: Kenechukwu Ogbuagu (manuscript), Sepehr Maleki (supervision and sections). All authors approved the final submitted draft.

\printbibliography

\end{document}